\documentclass[twocolumn]{wlscirep}
\usepackage[utf8]{inputenc}
\usepackage[T1]{fontenc}
\usepackage{indentfirst}
\usepackage{ragged2e}
\usepackage[font=small,labelfont=bf, skip=5pt, justification=justified, format=plain]{caption}
\usepackage{xspace}
\usepackage[style=science, backend=biber]{biblatex}
\AtBeginBibliography{\footnotesize}
\let\cite\autocite
\usepackage{times}
\usepackage{graphicx}
\usepackage[normalem]{ulem}
\usepackage[switch]{lineno} 
\usepackage{csquotes}

\usepackage{color}
\usepackage[dvipsnames]{xcolor}
\usepackage{hyperref}
\hypersetup{
	colorlinks,
	linkcolor={red!100!black},
	citecolor={blue!100!black},
	urlcolor={blue!80!black}
}
\usepackage{amssymb, amsmath, bm}
\usepackage{microtype}
\usepackage[font={footnotesize}]{caption}

\title{
Neural reservoir control of a soft bio-hybrid arm 
} 

\makeatletter
\renewcommand\AB@affilsepx{, \protect\Affilfont}
\makeatother

\author[1,5$\dag$]{Noel Naughton}
\author[2,5 $\ddagger$]{Arman Tekinalp}
\author[3,5 $^\S$]{Keshav Shivam}
\author[2,5]{Seung Hung Kim}
\author[3,5]{Volodymyr Kindratenko}
\author[2,3,4,5*]{Mattia Gazzola}
\affil[1]{\footnotesize{Beckman Institute for Advanced Science and Technology, University of Illinois Urbana-Champaign}}
\affil[2]{\footnotesize{Mechanical Science and Engineering, University of Illinois Urbana-Champaign} }
\affil[3]{\footnotesize{National Center for Supercomputing Applications, University of Illinois Urbana-Champaign}}
\affil[4]{\footnotesize{Carl R. Woese Institute for Genomic Biology, University of Illinois Urbana-Champaign}}
\affil[5]{\footnotesize{The Grainger College of Engineering, University of Illinois Urbana-Champaign}}

\date{}

\begin{abstract}
\vspace{-45pt}

A long-standing engineering problem, the control of soft robots is difficult because of their highly non-linear, heterogeneous, anisotropic, and distributed nature. Here, bridging engineering and biology, a neural reservoir is employed for the dynamic control of a bio-hybrid model arm made of multiple muscle-tendon groups enveloping an elastic spine. We show how the use of reservoirs facilitates simultaneous control and self-modeling across a set of challenging tasks, outperforming classic neural network approaches. Further, by implementing a spiking reservoir on neuromorphic hardware, energy efficiency is achieved, with nearly two-orders of magnitude improvement relative to standard CPUs, with implications for the on-board control of untethered, small-scale soft robots.
 
\vspace{-15pt}

\end{abstract}

\newcommand\blfootnote[1]{%
  \begingroup
  \renewcommand\thefootnote{}\footnote{#1}%
  \addtocounter{footnote}{-1}%
  \endgroup
}

\begin{document}
\flushbottom
\maketitle
\thispagestyle{empty}

\blfootnote{$^\dag$Current address: Department of Mechanical Engineering, Virginia Tech, Blacksburg, VA. $^\ddagger$Current address: Department of Mechanical Engineering, University of Maryland, College Park, MD. $^\S$Current address: Google, Mountain View, CA, $^*$email: mgazzola@illinois.edu}
Hyper-redundancy, underactuation, distributedness, and continuum mechanics are defining features of soft robots (artificial or biological \cite{Ricotti:2017,Morimoto:2018,Li:2025,Ren:2025,Park:2016,Pagan-Diaz:2018,Aydin:2019,Kim:2023}), intrinsic to their compliant, elastic constitutive materials. These traits are attractive in the pursuit of extreme reconfigurability, morphological adaptivity, delicacy and dexterity, for applications in medicine, defense, or agriculture \cite{Yasa:2023, Polygerinos:2017, Cianchetti:2018, Chowdhary:2019}. However, the same advantageous traits pose fundamental challenges in control, due to the associated vast space of degrees of freedom and the highly non-linear dynamics involved.

Recognizing the importance of the problem, significant effort has been devoted to the development of control approaches suitable for continuum and elastic structures \cite{Russo:2023,Della:2023, George:2018}. 
Model-based controllers have proven effective in quasi-static settings, but lack accuracy when inertial effects become significant and typically rely on simplifying assumptions that may overlook environmental interactions, anisotropy, or material nonlinearities \cite{Della:2023}. 
Data-driven approaches bypass such modeling difficulties by directly learning associated dynamics via artificial neural networks \cite{Chen:2024, Falotico:2024}. 
These techniques are generally resource-intensive and may employ neural architectures poorly suited to capture the long-range spatial and temporal dependencies that define soft systems \cite{Yasa:2023,Mengaldo:2022}, hindering robustness.

An alternative, bioinspired route to soft robotic control hinges on the concept of `mechanical intelligence' \cite{Pfeifer:2006, Ulrich:1988, Wang:2023_snake}, whereby non-linear structural and environmental effects are purposefully exploited to passively achieve adaptivity, thus simplifying and robustifying control. Complementarily, rhythm-generation techniques inspired by the nervous system (e.g., central pattern generators) have also been shown effective, particularly for locomotion \cite{Ijspeert:2008, Ramdya:2023}, leading to bio-hybrid attempts where neural-tissue is directly integrated with soft scaffolds, muscles, and electronics, to drive simple actuation dynamics \cite{Aydin:2019,Aydin:2020,Kim:2023,Tetsuka:2024}.

% \begin{figure*}[t!]
%     \centering
%     \includegraphics[width=\textwidth]{figures/Fig1.pdf}%
% 	\caption{
% 		 	\footnotesize{\textbf{Caption Description.} a) Schematic of neural reservoir control of a soft muscular arm. b) RL learning curves for control of a soft muscular arm tracking a moving target for an RC architecture vs. traditional feed-forward network architectures. c) RL learning curves for RC-based control for increasing reservoir sizes. d) Average episode scores for 180 episodes (1 hour) for different architectures. e) Side view of multiple snapshots overlaid. f) Top view of multiple snapshots overlaid. 			 	}}
%     \label{fig:1}
% \end{figure*}

\begin{figure*}[t!]
    \centering
    \includegraphics[width=\textwidth]{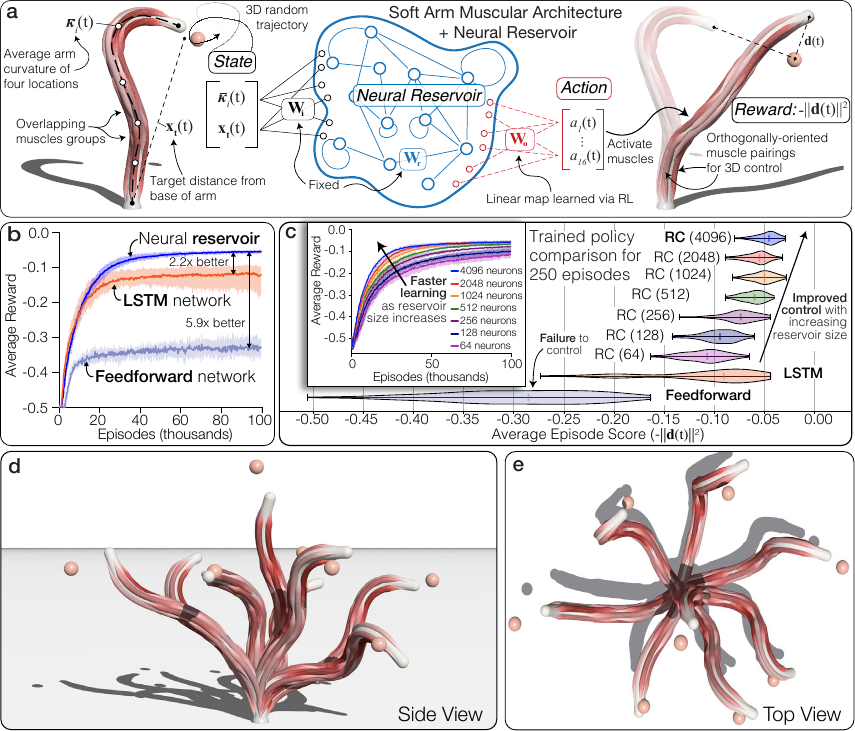}%
	\caption{
		 \footnotesize{\textbf{Neural reservoir control of a soft bio-hybrid} arm.} \textbf{(a)} Schematic depicting the coupling of a neural reservoir with a soft bio-hybrid arm made of sixteen muscle-tendon units enveloping an elastic spine to achieve autonomous control. \textbf{(b)} Learning performance for the control of a soft bio-hybrid arm (backbone stiffness: 125 kPa) tracking a 3D moving target: neural reservoir and traditional feedforward and LSTM network architectures. The solid lines depict the average learning performance (average episode return), each obtained by considering five neural architectures instances initialized using different random number generator seeds. The shaded regions depict the spread in learning performance  of the five architectures with different initial seeds. \textbf{(c)} Violin plots of average performance of trained policies evaluated over 250 episodes, for different reservoir sizes and reference architectures (FF and LSTM). Inset shows learning performance as reservoir size increases, illustrating how both more rapid and better overall learning performance is achieved as the reservoir size increases. For FF and LSTM, different numbers of neurons and layers / stacks were considered (FF: [64$\times$64], [128$\times$128], [64$\times$64$\times$64], [64$\times$64$\times$64$\times$64]; LSTM: [64$\times$1 stack], [128$\times$1 stack], [256$\times$1 stack], [64$\times$2 stacks], [128$\times$2 stacks], [256$\times$2 stacks]), with no significant differences observed. Here we report data for the best performing networks (SI for full details).
        Overlaid snapshots of \textbf{(d)} side view and \textbf{(e)} top view of the bio-hybrid arm successfully tracking a 3D target (see also SI Video 1) using a trained neural reservoir control policy. Intensity of muscle color (pale to dark red) denotes a muscle's activation level.}
    \label{fig:1}
    \vspace{-12pt}
\end{figure*}

Here, with the goal of intimately meshing compliant mechanics and neural dynamics for enhanced adaptivity and control, we numerically explore the integration of a soft bio-hybrid arm with Reservoir Computing (RC). Partially inspired by the mammalian neocortex \cite{Seoane:2019}, RC was developed \cite{Maass:2002} for processing real-time, analog, spatiotemporally correlated inputs, by means of a dynamically rich substrate \cite{Gonon:2019,Lukovsevivcius:2009,Maass:2005}, the reservoir. 
The reservoir, which in principle can be any dynamical system \cite{Tanaka:2019}, integrates and projects input data streams into a separable, high-dimensional latent space that decomposes non-linear correlations. 
Reservoir dynamics are then sampled and  recombined via linear maps into desired computations. While a variety of algorithmic and physical implementations have been proposed \cite{Lukovsevivcius:2009, Tanaka:2019}, in general the RC paradigm offers a number of appealing features \cite{Jaeger:2001, Maass:2002}: it does not require expensive backpropagation (only output linear maps are learned), it intrinsically accommodates nonlinear, spatiotemporally correlated dynamics, it is natively parallel as multiple maps can be learned separately to achieve different tasks while running on the same reservoir, and can be matched with specialized hardware (e.g., neuromorphic systems for energy efficiency \cite{Davies:2018, Merolla:2014}) or `wetware' (neural tissue used as bio-hybrid reservoir \cite{Zhang:2024}).

In this work, we  consider a neural reservoir (modeled as a recurrently connected or spiking network) able to sense the shape (proprioception) of a simulated bio-hybrid arm made of muscles and tendons enveloping an elastic spine. Neural dynamics are coupled to muscle actuations through Reinforcement Learning (RL), to learn to control the system's musculature in an unsupervised manner. We demonstrate that our approach produces control polices that outperform traditional deep-learning methodologies, with advantages seen to widen as the arm's compliance increases. RC’s high-dimensional latent space is also leveraged, through parallel output maps, for concurrent self-modeling to improve control robustness during conditions of disturbance and sensing failure. Further, motivated by the their potential relevance in bio-hybrid contexts, spiking neural reservoirs are considered, and subsequently mapped onto neuromorphic hardware, attaining a seventy-five fold reduction in energy usage. Neuromorphic RC is finally employed to drive the arm through a set of unstructured obstacles, learning to exploit solid objects to reshape and facilitate the reaching of a target --- a hallmark of mechanical intelligence that is automatically recovered here.

Overall, this work not only advances soft robotic control, but also furthers neuro-mechano integration, a characteristic trait of organic systems. This, in turn, may lead to novel bio-hybrid designs bridging engineering and biology \cite{Aydin:2019,Kim:2023,Tetsuka:2024,Zhang:2024, Cai:2023, Kagan:2022} as well as new insights into biological processing.

\textbf{Modeling of a soft bio-hybrid arm.} 
While no soft, synthetic actuator has yet reached biologically comparable levels of performance, improved artificial \cite{Duduta:2019,Wang:2021} or tissue-bioengineered \cite{Kim:2023} prototypes are continually being developed. Cognizant of this rapid development, we consider the automated, dynamic control of a computational neuromuscular arm, characterized by multiple, antagonistic and overlapping muscle-tendon pairs arranged around and along an elastic supporting spine (Fig.~\ref{fig:1}a).

Such a design captures key organizational motifs commonly encountered in biology and, increasingly, in robotics \cite{Wang:2023_snake, Ramdya:2023}. Antagonistic muscles are, for example, ubiquitous in both limbed \cite{Gray:1944} and limbless animals \cite{Kier:1985,Jayne:1988}, and  popular in bio-hybrid robots \cite{Ren:2025, Kim:2023, Morimoto:2018, Lee:2022}. 
Their asymmetric contractions enable rotations (i.e., joints \cite{Gray:1944}) or continuum bending and twisting (e.g., octopus arms \cite{Kier:1985, Tekinalp:2024}), while graded co-contractions can improve structural stability \cite{Gribble:2003,Koelewijn:2022} by actively increasing body stiffness. Similarly, overlapping muscle groups (of which the many staggered spinal muscles of snakes provide an extreme example \cite{Jayne:1988}) enable precise, distributed, and efficient actuation \cite{Zhang:2019}. Our prototypical model arm then allows us to explore compliant actuation and control in a biologically and engineering relevant setting, while offering a generalizable blueprint for experimental implementations. For example, the elastic spine may be realized at various scales via molding or 3D printing, and decorated with arrays of compliant actuators, from advanced supercoiled artificial muscles \cite{Spinks:2021} to cultured muscles \cite{Kim:2023, Wang:2021_biowalker}. 

To computationally instantiate our arm, we employ a modeling approach based on assemblies of Cosserat rods (Methods, \cite{Antman:2005,Gazzola:2018}). These are slender, one-dimensional elements able to undergo all modes of deformation ---bend, stretch, twist, and shear--- to dynamically reconfigure in three-dimensional space. Cosserat rods are a convenient representation since they naturally map to elastic beams (spine), tendons, and muscles, can actively contract and stretch along their length according to prescribed force-length relations, and can be connected together (via appropriate boundary conditions) into arbitrary architectures. They are thus well-suited to capture the heterogeneous, anisotropic, and distributed nature of our bio-hybrid arm. Assemblies of rods and their governing equations are numerically discretized and solved via the open-source software Elastica \cite{PyElastica}, whose quantitative utility has been demonstrated in a range of biophysical settings, including animal locomotion \cite{Zhang:2019, Zhang:2021, tekinalp2024soft}, invertebrate manipulation \cite{Tekinalp:2024}, plant dynamics \cite{Porat:2023}, fibrous metamaterials \cite{Bhosale:2022, Weiner:2020}, and soft robotic design and control \cite{Naughton:2021, Chang:2020,Chang:2023, Wang:2022, Shih:2023, Charles:2023, Zhang:2019, Kim:2023, Wang:2021_biowalker}.

We begin with the soft elastic spine/backbone, represented as a single passive rod, onto which 16 muscle-tendon units are patterned (Fig.~\ref{fig:1}a, Fig.~S1). Three-dimensional arm deformations are achieved by organizing muscle-tendon units into four layers of orthogonal agonist-antagonistic pairs, with adjoining layers overlapping by 50\% of their length and offset 45$^\circ$ to avoid intersection. This architecture enables bending in all directions, via the coordinated activation of orthogonal muscles combined with the omni-directional bending/twisting capability of the elastic backbone. Further, muscle overlap allows for continuously graded contractions along the arm despite the limited number of individual muscles \cite{Zhang:2019}. Each muscle-tendon unit is modeled as an individual rod with an actively-contractible muscle belly and tapered tendinous ends that insert into the backbone. The entire unit is glued lengthwise to the spine, to conform to it while bending, mimicking the presence of surrounding fascia tissue or encapsulating materials.

The active force-length and force-velocity relationships of the muscle belly and its passive hyperelastic material properties as well as the tendon's material properties are based on reported biomechanical data (Methods). Noteworthy is the nonlinear passive stress-strain behavior of muscle and tendon tissues \cite{Calvo:2010, Maganaris:1999}, which are characterized by a `J-curve' response that allows for compliant deformations at small strains, before stiffening at larger strains for structural stability \cite{Ker:2007}. This is a key mechanical feature of biological musculoskeletal tissues, and significant engineering effort has been recently devoted to recapitulate similar properties \cite{Vatankhah:2017, Tu:2021, Lim:2019}. Finally, each muscle is independently controllable via a continuous tetanic activation ($a_i\in[0,1]$, $i\in\{1,...~16\}$). Full details are provided in Methods.

We challenge the arm to learn to coordinate its muscle activations (action space $\textbf{a}_t(t)$) to dynamically track a target moving along a smoothly-varying, random 3D trajectory (Fig. \ref{fig:1}a). First, we outfit the arm with a compact set of sensory capabilities. Analogous to strain-sensing muscle spindles in vertebrates \cite{Matthews:1964}, proprioception is provided here through the arm’s curvature $\bm{\bar{\kappa}}_i(t)$, estimated at four equally spaced locations (Methods). Environmental information is limited to the distance vector $\textbf{x}_t(t)$ between the (fixed) base of the arm and the moving target, representing visual or acoustic tracking. Notably, this sensory arrangement is laboratory-frame invariant. Indeed, local curvature is sufficient to recover the arm shape, while the target is sensed relative to the arm base. Thus, overall, the arm state is constituted by $\bm{\bar{\kappa}}_i(t)$ and $\textbf{x}_t(t)$, with all higher-order information (velocities or accelerations) implicitly captured through the memory of the reservoir \cite{Lukovsevivcius:2009}, described next.

\textbf{Neural reservoir control.}
Controlling this structure is challenging because of the arm's continuum elastic nature, characterized by highly non-linear and long-range stress propagation effects. Indeed, distributed and localized loads (musculature/environment) are communicated throughout the entire system, potentially leading to mechanical instabilities and global morphological reconfigurations. Compounding this complexity are the non-linear contractile properties of the muscle-tendon units, the anisotropic and heterogeneous quality of the arm, as well as its inertial dynamics.

To establish control (Fig.~\ref{fig:1}a), we seek to learn, in an unsupervised, model-free  fashion, muscle activations (action: $a = \{a_j(t)\}$, $j\in\{1,...16\}$) that minimize at all times the distance $\mathbf{d}(t)$ between the arm tip and the target (reward: $r=-||\mathbf{d}(t)||^2$), based on available sensory information (state: $s = \{\textbf{x}_t(t),~\bm{\bar{\kappa}}_i(t)\}$, $i\in\{1,...4\}$). The mapping (control policy) between state and action space may be represented by any neural network topology \cite{Hagan:1999,Perrusquia:2021}, and it is here that we insert our neural reservoir (Fig.~\ref{fig:1}a). This is initially realized as a recurrent artificial neural network (Methods) paired with a single, external linear recombination layer (map) whose function is to transform the reservoir’s internal dynamics (sampled at each neuron) into muscle contractions $a = \{a_j(t)\}$. To learn this linear map, we employ the reinforcement learning (on-policy) proximal policy optimization (PPO) strategy \cite{Schulman:2017}. We emphasize here that since RC requires learning the output map weights only, all connections from the input state to the reservoir and within the reservoir itself, remain fixed, removing the need for backpropagation.

\begin{figure*}[t!]
    \centering
    \includegraphics[width=\textwidth]{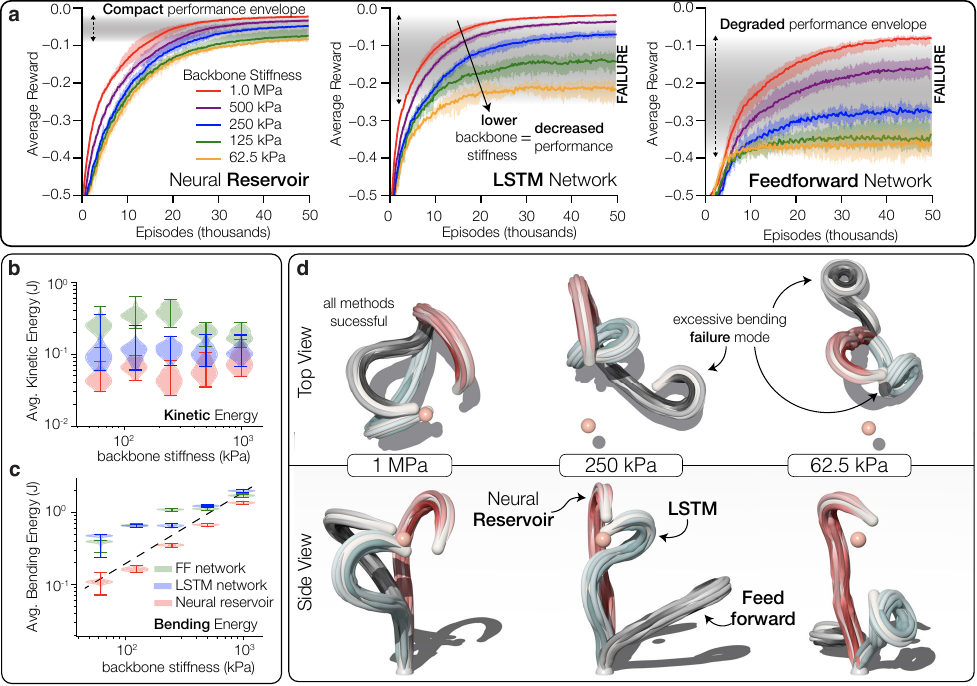}%
    \caption{ \textbf{Control of increasingly compliant bio-hybrid arms}. 
    \textbf{(a)} Learning performance of a neural reservoir (4096 neurons) compared to LSTM and FF networks for a soft bio-hybrid arm with decreasing backbone elastic modulii. The neural reservoir exhibits a compact performance envelope while the LSTM and FF networks exhibit much wider performance envelopes with control performance drastically decreasing as the backbone softens. Solid and shaded regions denote the performance average and spread relative to five randomly initialized network instances (as in Fig.~\ref{fig:1}b). 
    Violin plots of the average \textbf{(b)} kinetic and  \textbf{(c)} bending energy of the arm's spine, when RC, LSTM, and FF network architectures are used for controlling systems of decreasing backbone stiffness. Values reported are total kinetic/bending energy integrated over the episode and normalized by episode length. The RC network exhibits lower energies in all cases. Bending energy is observed to decrease linearly in the case of RC, in keep with the linear softening of the spine and in contrast with FF and LSTM, which exhibits sublinear scalings. 
    \textbf{(d)} Snapshots of trained policy performance for different backbone stiffness levels illustrating control failure modes. While all three network architectures successfully control a soft arm with a backbone stiffness of 1 MPa, for a stiffness of 250 kPa the feedforward network fails to coordinate muscle contractions, demonstrating an excessive bending failure mode (with an associated increased bending energy compared to RC, see panel (b)). While The LSTM network exhibits better performance, it also produces  excessive bending and fails as the backbone stiffness continues to decrease (62.5 kPa). Video comparison of tracking performances for all backbone stiffness cases is available in SI Video 2. As in Fig.~\ref{fig:1}, for FF and LSTM, different numbers of neurons and layers / stacks were considered, with no significant differences observed across arm's stiffnesses. Here we report data for the best performing networks (SI for full details).
    }
    \label{fig:2}
    \vspace{-12pt}
\end{figure*}

To contextualize the reservoir performance, we additionally consider feedforward (FF) and long short-term memory (LSTM) networks, reference architectures commonly employed in learning-based control \cite{Yu:2019,Zou:2009}. Training is again performed using PPO, where this time the network weights are all learned (as opposed to RC) through backpropagation. To facilitate a meaningful comparison, all network topologies are evaluated against the same control problem, and setup/trained according to the same protocol (Methods). 

As illustrated in Fig.~\ref{fig:1}b-e and SI Video 1, upon training, the neural reservoir successfully learns to track the target, substantially outperforming the FF and LSTM networks. Indeed, the FF network is found to quickly reach a (poor) performance limit beyond which it is unable to improve (Fig.~\ref{fig:1}b). The LSTM network instead converges to a better performance, although the ceiling remains significantly below the reservoir's (which is 2.2x better). The neural reservoir's performance is also found to correlate with reservoir size, whereby learning speed and tracking accuracy increase as the number of neurons grows (Fig.~\ref{fig:1}c). This is in keeping with the known ability of larger reservoirs to capture complex dynamics \cite{Pathak:2018,Lukovsevivcius:2009}, and  supports the hypothesis that the control of elastic systems may benefit from a specific focus on highly non-linear, spatiotemporally correlated dynamics.

To further test this hypothesis, we investigate the effect of the arm's elastic spine stiffness by progressively decreasing its elastic modulus (while holding mass and geometry constant). This softening enables larger modes of deformations, further exacerbating the system’s non-linear response, rendering control harder and allowing us to tease out differences between RC, FF and LSTM approaches. We begin with a relatively stiff backbone (1 MPa, equivalent to rubber \cite{Nagdi:1993} and $\sim$10x stiffer than Fig. \ref{fig:1}b) for which all three network topologies successfully learn to track the moving target (Fig.~\ref{fig:2}a, SI Video 2). However, as soon as the spine softens, differences start to manifest. The learned FF controllers are indeed observed to rapidly degrade, practically failing to coordinate the arm already at 500 kPa (SI Video 2). The LSTM network instead initially maintains control, although below 250 kPa its performance too rapidly degrades (SI Video 2). In contrast, the neural reservoir exhibits a compact performance envelope and maintains sustained control down to a 62.5 kPa stiffness (comparable to mammalian \cite{Fish:1984} or octopus \cite{DiClemente:2021} muscle tissue \cite{Fish:1984}, for reference). 

\begin{figure*}[t!]
    \centering
    \includegraphics[width=\textwidth]{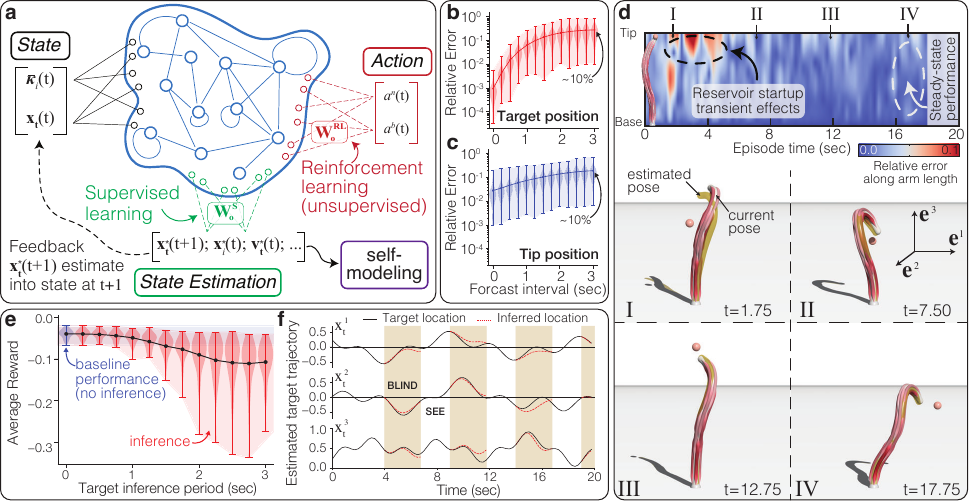}%
	\caption{
        \footnotesize{
            \textbf{Parallel maps for self-modeling and robust control.} 
            \textbf{(a)} Schematic of neural reservoir control equipped with additional, parallel maps to infer/predict state information. All reported results are for an arm with a backbone stiffness of 250 kPa controlled by an already trained (via RL) neural reservoir with 4096 neurons, as described in Fig.~\ref{fig:1}. 
            \textbf{(b)} Accuracy of parallel map estimates of future target positions, for increasing time-windows into the future. 
            \textbf{(c)} Accuracy of parallel map estimates of future arm's tip position for increasing time-windows into the future. For both (b) and (c), the relative error is defined as $ ||\mathbf{\hat{y}} - \mathbf{y}||/\ell$ where $|| \cdot ||$ is the L2-vector norm, $\mathbf{\hat{y}}$ is the predicted position, $\mathbf{y}$ is the true position, and $\ell$ is the length of the soft arm. 
            \textbf{(d)} Performance of reservoir self-modeling for estimation of current arm pose (top row). Heat map of accuracy of pose estimation along the length of the arm. Color denotes relative error of estimation of the position of a point $s\in[0,\ell]$ along the soft arm of length $\ell$. Relative error is defined as $||\mathbf{\hat{y}}(s) - \mathbf{y}(s)||/s$ where $\mathbf{\hat{y}}(s)$ is the predicted location of point $s$ and $\mathbf{y}$(s) is the true location of that point. Accuracy is initially lower due to transient startup effects in the reservoir before reaching a consistent level of high accuracy as confirmed (mid/bottom row) by visualization of the estimated (gold) and true (full color) arm pose at selected time instances}. 
            \textbf{(e)} Tracking performance of the neural reservoir when the target position becomes unavailable for increasing lengths of time. The target oscillates between being measured (`seen') and inferred (`blind') for equal time intervals that range in length from 0.25 seconds to 3 seconds. Violin plots show performance over 50 trials for increasing periods of time during which the arm is blinded. The blue shaded region shows the baseline performance of the reservoir when the target location is always known.
            \textbf{(f)} Comparison of inferred target's 3D position for a three-second inference period. After approximately 1.5 seconds, the estimate of the target trajectory sometimes drifts from the true trajectory, leading to the drop in tracking performance seen in panel (e).
        }
    \label{fig:3}
    \vspace{-12pt}
\end{figure*}

An energetic analysis reveals that the neural reservoir consistently minimizes (relative to FF and LSTM, Fig.~\ref{fig:2}b,c) both the kinetic and bending energy of the spine. For each network architecture, kinetic energy (Fig.~\ref{fig:2}b) is found to remain approximately consistent across Youngs' moduli, albeit settling at different levels, with LSTM and FF entailing $\sim$3x and $\sim$10x larger energies than RC, respectively. Phenomenologically, higher kinetic energies manifest as faster swings or pronounced structural vibrations (SI Video 2), and thus the RC-based controller (which produces reduced kinetic energies) exhibits more precise and smoother actuation, without extensively relying on  corrective muscle contractions. The neural reservoir also demonstrates clear superiority in minimizing the spine’s bending energy (Fig.~\ref{fig:2}c). It is worth noting that bending stiffness scales linearly with the elastic modulus. We would thus expect that a controller able to effectively coordinate the arm would produce bending deformations (and therefore energies) congruent with this linear scaling as the spine softens. This trend is indeed met by RC, while LSTM and FF networks are found to exhibit a sub-linear decrease. This indicates that LSTM and FF networks resort to deformations that are excessive in relation to the spine’s stiffness, eventually leading to failure (SI Video 2). Recorded muscle activation patterns (SI Fig. S2) suggest that RC's greater control may be enabled by the use of antagonistic muscle co-contractions, to locally stiffen and stabilize the arm as it softens. LSTM and FF controllers instead do not discover this strategy, and forgo co-contractions in favor of one-sided activations.

Therefore, RC-based controllers are shown to achieve higher tracking accuracies and, concurrently, avoid unnecessary bending, twisting, or corrective accelerations, reflecting their ability to capture (and predict) the arm’s dynamics, as quantified next.

\begin{figure*}[t!]
    \centering
    % \vspace{15pt}
    \includegraphics[width=\linewidth]{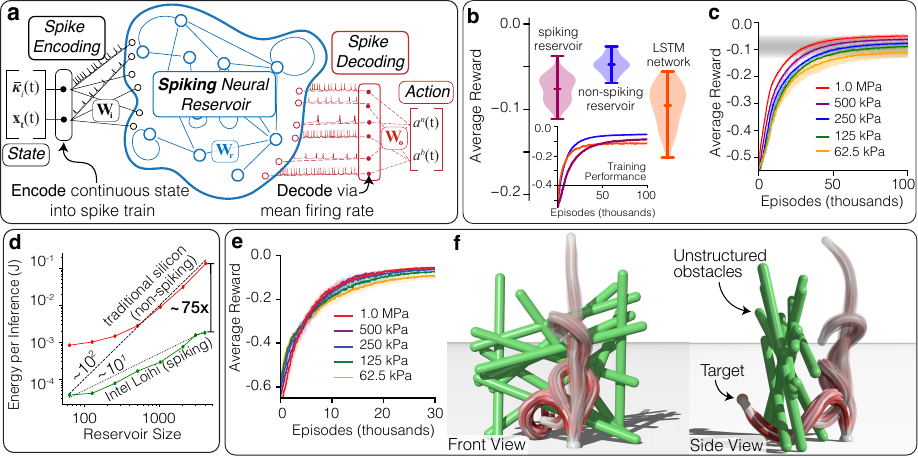}%
    \caption{ \textbf{Spiking neural reservoir control on neuromorphic hardware} \textbf{(a)} Schematic showing how the state is encoded into a spike train (SI) that is sent to a spiking neural reservoir running on an Intel Loihi chip. Spike train outputs are then decoded (SI) into continuous actions (contractions of the arm muscles).
    % The resulting spike trains are then decoded to continuous actions based on their mean firing weight and learned output weights. 
    \textbf{(b)} Comparison between the performance over 250 episodes of a trained spiking neural reservoir on Loihi (backbone stiffness: 125 kPa; reservoir size: 2048 neurons, which is the largest implementable on Loihi), a neural reservoir running on traditional silicon using non-spiking artificial neurons, and an LSTM network (same as Fig.~\ref{fig:1} and Fig.~\ref{fig:2}). Inset shows the training performance for the spiking reservoir, the non-spiking reservoir, and the LSTM.
    % : Performance of final trained policies with the spiking network running on the Intel Loihi neuromorphic chip and other networks running on an x86 CPU. The spiking reservoir exhibits similar performance than the non-spiking RC and outperforms the LSTM and feed-forward approaches. 
    \textbf{(c)} The spiking reservoir exhibits a compact performance envelope as the backbone stiffness decreases. All cases are trained using five random initialization seeds as in Fig.~\ref{fig:1}b.
    % \textbf{(d)} Control performance increases as the size of the spiking reservoir increases. Due to limitations of the Loihi chip, a maximum reservoir size of 2048 neurons was considered. 
    \textbf{(d)} Energy use of a non-spiking reservoir running on traditional silicon (Intel Xeon W-2665) and the spiking reservoir running on the Intel Loihi chip. The non-spiking reservoir on traditional silicon exhibits quadratic energy scaling as the reservoir size increases compared to the linear energy scaling of the spiking reservoir running on the Intel Loihi chip.
    \textbf{(e)} Initial 30k episodes of learning performance of neural reservoirs (n=1024) controlling arms of decreasing backbone stiffness tasked with reaching through a cluttered nest of obstacles to a fixed target (SI for expanded results). Video of the performance of final trained policies for all stiffness levels is available in SI Video 6.
    \textbf{(f)} Timelapse front/side views of a spiking reservoir guiding a soft arm through  unstructured obstacles to reach a target (backbone stiffness: 125 kPa).}
    \label{fig:4}
    \vspace{-12pt}
\end{figure*}

\textbf{Self-modeling for robust control.} 
Reservoir dynamics can also be used for self-modeling \cite{Chen:2022, Hu:2025}, that is, for the explicit estimation or prediction of the system’s configuration, which in turn can be used to robustify the controller.
Self-modeling is achieved here through three dedicated maps (Fig.~\ref{fig:3}a), to extract the arm's full pose, future target location, and future tip location. Maps are obtained in a supervised fashion via Ridge regression based on data from the 250kPa case of Fig.~\ref{fig:2}a (Methods). The maps are then directly appended to the reservoir, exploiting RC’s intrinsic parallelism whereby several computations can be carried out concurrently using the same underlying dynamics.

We show how the reservoir, using target location and curvature proprioception history, accurately synthesizes future target locations (Fig.~\ref{fig:3}b), current arm pose (Fig.~\ref{fig:3}d, SI Video 3), and future tip positions (by anticipating forthcoming muscle activations selected by the control policy based on the predicted target trajectory, Fig.~\ref{fig:3}c). We note that in all cases target trajectories are randomly generated so that the reservoir must constantly produce new predictions. These capabilities are then directly used to improve control in a scenario where the arm intermittently loses sight of the target, mimicking a faulty sensor or environmental disturbances. For the arm to keep `seeing', the target's next location is predicted and fed back into the reservoir in lieu of the missing data. This in turn enables the forecasting of muscle actuations and future tip positions, allowing the blinded arm to continue tracking (Fig.~\ref{fig:3}e,f). Using this strategy, inference periods of less than one second (interval during which the arm is `blind', approximately equivalent to the target traveling a distance of 50\% the arm’s length) result in no performance degradation relative to perfect knowledge of the target location. If the disturbance lasts beyond one second, then tracking quality begins to decrease, although modestly. Indeed, average performance remain comparable to an LSTM network with complete knowledge of the target location (SI). However, the spread of the arm performance (violin plot bars in Fig.~\ref{fig:3}e) increases for large disturbance periods, implying that the arm does occasionally lose track of the target while blinded.

Thus, a mixed strategy where unsupervised learning is used for control and supervised learning is employed for self-modeling, with the second supporting the first, is found to be effective in robustifying performance (SI Video 4). It is worth noting that the inclusion of self-modeling entails minimal computational overhead since all linear maps use the same reservoir.

\textbf{Spiking reservoirs on neuromorphic hardware.}
Here, we illustrate how our approach is also well-positioned to enhance computational energy efficiency, a concern in many robotic applications \cite{Zhu:2024,Sandamirskaya:2022}. To this end, we consider a spiking neural reservoir on neuromorphic hardware, demonstrating almost two orders of magnitude improvements relative to standard CPUs, while retaining control performance.

The use of spiking networks, inspired by cortical dynamics \cite{Lukovsevivcius:2009, Dominey:1995}, well-aligns with both RC \cite{Maass:2002} and specialized neuromorphic chips \cite{Davies:2018, Merolla:2014} that physically implement energy efficient artificial spiking neurons \cite{Rajendran:2019, Javanshir:2022}. It is thus only natural to explore the use of spiking networks within our framework, also envisioning potential bio-hybrid applications where spiking neural tissue may replace synthetic reservoirs.

To match algorithmic and hardware infrastructures, we consider a reservoir of leaky integrate and fire (LIF) spiking neurons on Intel's neuromorphic Loihi chip \cite{Davies:2018} (Fig.~\ref{fig:4}a, Methods). This spiking reservoir is challenged to control our soft arm and is compared with the LSTM and the non-spiking reservoir approaches of Fig.~\ref{fig:1}, which were run on traditional CPUs (Intel Xeon).

As can be seen in Fig. \ref{fig:4}b, both reservoir-based approaches outperform the LSTM network in the target-tracking test. While the non-spiking reservoir exhibits slightly better overall performance, inspection of the trained policies (SI Videos 1 \& 5) shows that both approaches are in fact successful in controlling the soft arm, which is instead found to frequently collapse (as captured by the larger score spread) in the LSTM case. Further, consistent with the non-spiking reservoir, the spiking reservoir exhibits a compact performance envelope as the arm softens (Fig.~\ref{fig:4}c), and larger reservoirs improve performance (SI Fig. S4).

However, the spiking reservoir on the neuromorphic Loihi chip provides a clear advantage when energy use is considered (Fig.~\ref{fig:4}d). Indeed, power consumption is observed to reduce by up to 75x. Furthermore, and critically, energy use on Loihi exhibits linear scaling with reservoir size compared to quadratic scaling for reservoirs deployed on traditional silicon hardware. This implies that larger (thus more capable) spiking reservoirs may be used in energy-constrained applications.

\textbf{Environmental interactions.}
A key advantage of soft robots is their ability to comply with the environment without damaging themselves or their surroundings. We explore these boundary interactions, and the possibility of taking advantage of them, by considering the reaching of a target through a set of unstructured obstacles (Fig.~\ref{fig:4}e,f). We underscore the steep control challenge associated with this test, exacerbated by a complex environment, geometric constraints, and contact dynamics.

In Fig.~\ref{fig:4}e,f, we show results obtained using a spiking neural reservoir for arms characterized by spines of decreasing stiffness. As can be seen, all arms successfully learn to reach through the obstacles with softer ones exhibiting faster initial learning, ostensibly due to their ability to better conform to obstacles and push through gaps without precise actuation. However, the more precise control associated with stiff arms is found to eventually achieve better performance (Fig.~\ref{fig:4}e). In all cases though, throughout training, arms are observed to extensively exploit (instead of avoiding) solid boundaries, leaning against them to passively reconfigure and favorably redirect the tip (SI Video 6). This facilitates learning and control, as supported by the fact that the target can be reached even by the softest arm (see Video), despite being the most difficult to coordinate (Fig.~\ref{fig:2}). Thus, our RC approach is found to naturally recover a hallmark of mechanical intelligence, whereby passive mechanics and the environment are leveraged to simplify the task at hand.

\textbf{Discussion.}
In this work, we consider the challenging problem of learning to control a heterogeneous, soft bio-hybrid arm. The arm’s musculoskeletal architecture is designed to encompass a range of biomechanically relevant elements and is modeled via assemblies of Cosserat rods. The physical infrastructure is matched with a neural infrastructure (the reservoir), to encode  the arm’s sensing and spatiotemporal dynamics. This representation is coupled with reinforcement learning to achieve muscle coordination and allow the arm to track and reach targets.

Our approach is demonstrated to be effective, outperforming classic neural network methods. In particular, the synergy between neural and physical dynamics is observed to produce control policies that minimize (relative to FF and LSTM) the arm’s bending and kinetic energies, across a range of material stiffnesses. In contrast, FF and LSTM approaches are found unable to cope with material property changes. Parallel self-modeling on the same reservoir is also demonstrated for robust control, in the case of unavailable or corrupted sensory information. Further, spiking neural reservoirs on neuromorphic hardware are shown to deliver over 75-fold energy savings relative to non-spiking reservoirs, while conserving control performance. Spiking reservoirs are finally employed to drive the arm through a set of unstructured obstacles, taking advantage of solid boundaries and compliance, a hallmark of mechanical intelligence.

The capacity of neural reservoirs to synthesize compliant dynamics presents much potential for soft robotic control, combining learning, versatility, adaptivity, and robustness to mechanical changes or environmental disturbances. Higher-order situational information (velocities or accelerations) can be extracted with minimal additional computational costs (parallel maps running on the same reservoir), reducing onboard sensing and energy requirements. This aspect is compounded by the use of spiking neural reservoirs on neuromorphic hardware, to deliver significant energy savings, a necessity in many robotic applications. 

These results not only advance soft robotic control, but may also further the use of biological constructs in bio-hybrid robotics and engineering, with both spike-activated cultured muscle tissue \cite{Pagan-Diaz:2018, Aydin:2019, Wang:2021_biowalker, Kim:2023} and living reservoirs of cultured neural tissue \cite{Zhang:2024} amenable to integration within this computing paradigm.

%% bib
{\footnotesize
\printbibliography
}

{\footnotesize
\vspace{-10pt}
\subsection*{Acknowledgements}

\noindent
\textbf{Funding:} This study was jointly funded by ONR MURI N00014-19-1-2373 (M.G.), ONR N00014-22-1-2569 (M.G.), NSF EFRI C3 SoRo \#1830881 (M.G.), NSF CAREER \#1846752 (M.G.), NSF Expeditions in Computing `Mind in Vitro' \#2123781 (M.G.), and NSF EFRI BEGIN OI \#2422340 (N.N.). Computational support was provided by the Bridges2 supercomputer at the Pittsburgh Supercomputing Center through allocation TG-MCB190004 from the Extreme Science and Engineering Discovery Environment (XSEDE; NSF grant ACI-1548562).
\textbf{Author contributions statement:}
N.N., A.T., K.S., and M.G. conceptualized and designed the research. N.N., A.T., K.S., and S.H.K. performed the research. All analyzed the data. All wrote the paper.
\textbf{Competing interests:}
The authors have no competing interests to declare.
% \textbf{Code availability:}
% Code for the setup and simulation of the muscular arm model presented in this paper is available in a supplementary attachment and will be made available on GitHub upon publication. 
% The open-source Python implementation of Elastica used in this paper is available at \url{www.github.com/GazzolaLab/PyElastica}. 
% Code for the setup and simulation of the octopus arm model presented in this paper is available on GitHub (\href{https://github.com/GazzolaLab/topology-dynamics-control-of-octopus}{github.com/GazzolaLab/topology-dynamics-control-of-octopus}).

\noindent 

}
\section*{Methods}
\subsection*{Modeling a neuromuscular arm}

We model the muscular arm as an assembly of Cosserat rods, which are slender, one-dimensional elastic structures that can undergo all modes of deformation: bending, twisting, stretching, and shearing. Representing the backbone and muscle groups as Cosserat rods entails a number of advantages. Assembled together they naturally capture the heterogeneity of muscular architectures while the Cosserat formulation can be extended to account for non-linear material properties, connectivity, active stresses, internal pressure, and environmental loads. The numerical implementation of Cosserat rods is computationally efficient as they accurately capture large 3D deformations through a one-dimensional representation, alleviating time-consuming remeshing difficulties and compute costs of 3D elasticity.

% \vspace{5pt}
% \textit{\uline{Dynamics of a single Cosserat rod.}} 
\subsubsection*{Dynamics of a single Cosserat rod.} 
An individual Cosserat rod is described by its center line position $\mathbf{\bar{x}}(s,t) \in \mathbb{R}^{3}$ and an oriented reference frame of (row-wise) orthonormal directors $\mathbf{Q}(s,t) \in \mathbb{R}^{3 \times 3} =\left[\mathbf{\bar{d}}_{1}, \mathbf{\bar{d}}_{2}, \mathbf{\bar{d}}_{3}\right]^{-1}$ along its length $s \in [0, L(t)]$, where $L(t)$ is the current length, for all time $t \in \mathbb{R} \ge 0$. 
Any vector defined in the global \textit{lab frame} ($\mathbf{\bar{v}}$) can be transformed into the \textit{local frame} ($\mathbf{{v}}$) via $\mathbf{v} = \mathbf{Q}\mathbf{\bar{v}}$ and from the local to the lab frame via $\mathbf{\bar{v}} = \mathbf{Q}^{T}\mathbf{v} $. 
For an unshearable and inextensible rod, $\mathbf{\bar{d}}_{3}$ is parallel to the rod's local tangent ($\partial_s \mathbf{\bar{x}} = \mathbf{\bar{x}}_{s}$), and 
 $\mathbf{\bar{d}}_{1}$ (normal) and $\mathbf{\bar{d}}_{2}$ (binormal) span the rod's cross-section.
However, under shear or extension, the rod's tangent direction $\mathbf{\bar{x}}_{s}$ and $\mathbf{\bar{d}}_{3}$ are no longer the same, with the difference represented by the shear strain vector $\bm{\epsilon} = \mathbf{Q} \left(\mathbf{\bar{x}}_{s} -  \mathbf{\bar{d}}_{3} \right)$. 
The gradient of the directors ($\mathbf{\bar{d}}_j$) with respect to the rod's length is defined by the curvature vector $\bm{\bar{\kappa}}(s,t) \in \mathbb{R}^{3}$ through the relation $\partial_s \mathbf{\bar{d}}_{j} = \bm{\bar{\kappa}} \times \mathbf{\bar{d}}_{j}$. In the local frame ($\bm{\kappa} = \mathbf{Q} \bm{\bar{\kappa}}$), the components of the curvature vector $\bm{\kappa} = [\kappa_1,~\kappa_2,~\kappa_3]$ relate to bending (${\kappa}_{1}$ and ${\kappa}_{2}$) and twisting (${\kappa}_{3}$) of the rod.
Similarly, the gradient of the directors with respect to time is defined by the angular velocity vector $\bm{\bar{\omega}}(s,t) \in \mathbb{R}^{3}$ through the relation $\partial_t \mathbf{\bar{d}}_{j} = \bm{\bar{\omega}} \times \mathbf{\bar{d}}_{j}$. The linear velocity of the center-line is $\mathbf{\bar{v}}(s,t) \in \mathbb{R}^{3}=\partial_t \mathbf{\bar{x}}$ while the second area moment of inertia $\mathbf{I}(s,t)\in \mathbb{R}^{3 \times 3}$, cross-sectional area ${A}(s,t)\in \mathbb{R}$, and density $\rho(s)\in \mathbb{R}$ are defined based on the rod's material properties.
The dynamics of a Cosserat rod are then described as \cite{Gazzola:2018} 
\begin{equation} 
    \partial_{t}^{2} \left( \rho A \mathbf{\bar{x}} \right)= \partial_s \left(\mathbf{Q}^{T} \mathbf{n}\right) + \mathbf{\bar{f}}
    \label{eqn:linear_momentum}
\end{equation}
\begin{equation}\label{eqn:angular_momentum}
    \partial_t \left( \rho \mathbf{{I}} \bm{\omega} \right) = \partial_s \bm{\tau} + \bm{\kappa} \times \bm{\tau} + \left(\mathbf{Q} \mathbf{\bar{x}}_s \times \mathbf{n}\right)
     + \left(\rho \mathbf{I} \bm{\omega}\right) \times \bm{\omega} +  \mathbf{Q}\mathbf{\bar{c}}
\end{equation}
where Eq.~\ref{eqn:linear_momentum} (lab frame) and  Eq.~\ref{eqn:angular_momentum} (local frame) represent the change of linear and angular momentum at every cross-section, respectively,
$\mathbf{n}(s,t)\in \mathbb{R}^{3}$ and $\bm{\tau}(s,t)\in \mathbb{R}^{3}$ are internal forces and couples, respectively, developed due to elastic deformations and muscle contractions while $\mathbf{\bar{f}}(s,t)\in \mathbb{R}^{3}$ and $\mathbf{\bar{c}}(s,t)\in \mathbb{R}^{3}$ capture external forces and couples applied to the arm, respectively.

% \textit{\uline{Numerical solution.}}
The above continuous representation is discretized into $\left(n_{\text{elem}}+1\right)$ nodes of position $\mathbf{\bar{x}}_{i}$ that are connected by $n_{\text{elem}}$ cylindrical elements. Linear displacements are determined by the internal and external forces acting at the nodes, while rotations are accounted for via couples applied to the cylindrical elements. Energy losses due to internal friction and viscoelastic effects are captured through use of Rayleigh potentials.
The dynamic behavior of a rod is then computed by integrating the discretized set of equations, along with appropriate boundary conditions, in time via a second-order position Verlet scheme \cite{Gazzola:2018}. In this study, we used PyElastica \cite{PyElastica}, which is an open-source, Python-based implementation of this numerical scheme.

\subsubsection*{Assembling a muscular arm.}  
We assemble our muscular arm by arranging active and passive rods into a representative muscular architecture. The arm consists of an elastic backbone (modeled as a single rod) onto which sixteen muscle-tendon units (each modeled as a single rod) are attached around the outside for a total of seventeen rods. This mechanical connectivity enables the arm to translate the one-dimensional internal contraction forces generated by individual muscles into the three-dimensional dynamic motions of the arm as a whole. A description of the arm's geometry and material parameters is provided in the SI. 

\textit{\uline{Modeling the elastic backbone.}}
The backbone is modeled as a constant cross-section, elastic rod.
For a linear elastic material such as the backbone, the internal forces $\mathbf{{n}} = [n_1,~n_2,~n_3]$ are proportional to the shear strain of the rod $\mathbf{{n}} = \mathbf{{S}} \bm{\epsilon}$ where $\mathbf{S} = \text{diag}\left(\alpha_c G A, ~\alpha_c G A, ~E A \right)$ is the rod's shear/stretch stiffness matrix, $E$ is the rod's Young's modulus, $G$ is the rod's shear modulus, and $\alpha_c$ is the Timoshenko shear correction factor \cite{Gazzola:2018}. We additionally model internal elastic torques $\bm{\tau}$ using the linear relationship $\bm{\tau} = \mathbf{B} \bm{\kappa} $
where $\mathbf{B} \in \mathbb{R}^{3 \times 3}  = \text{diag}\left(E {I}_1, ~E {I}_2, ~G {I}_3 \right)$ is the rod's bend/twist stiffness matrix, and ${I}_{1}$, ${I}_{2}$, and ${I}_{3}$ are the rod's second moments of inertia about $\mathbf{\bar{d}}_1$, $\mathbf{\bar{d}}_2$, and $\mathbf{\bar{d}}_{3}$, respectively.
Finally, zero-displacement and zero-rotation boundary conditions are defined on the node and element of the backbone nearest to the base, respectively, to anchor the arm.

\textit{\uline{Modeling the muscle-tendon unit.}}
We model muscle-tendon units based on frog semitendionous muscle, in line with previous bio-hybrid demonstrations \cite{Herr:2004}.
Biological muscles actively generate internal forces that cause them to axially contract while both muscle and tendon tissues exhibit hyperelastic passive behavior when stretched. Both effects render a linear stress-strain treatment of the Cosserat rod's axial stretch inaccurate. Instead, we model the axial component ($n_3$) of the internal force vector $\mathbf{n}$ by considering the active $\sigma_{a}(s,t) \in \mathbb{R}$ and passive $\sigma_{p}(s,t) \in \mathbb{R}$ axial stresses developed along the muscle $n_3 = A (\sigma_a^m + \sigma_p^m)$ and the passive response of the tendon $n_3 = A (\sigma_p^t)$ while modeling its shear components ($n_1$, $n_2$) using the above presented linear elastic formulation. Internal elastic torques are represented using a similar approach as the elastic backbone, though we note that the non-linear effects of the muscle's axial stretch extends to encompass angular momentum via the third term in Eq. \ref{eqn:angular_momentum}. 
Material properties (provided in the SI) of the muscle-tendon unit are based on the sliding-filament model of muscle \cite{VanLeeuwen:1991, VanLeeuwen:1997} and experimental measurements \cite{Lieber:1992}. 
% are based on longitudinal muscles of squid tentacles \cite{VanLeeuwen:1997} while the tendon's material properties are based on a model of frog semitendinosus tendon \cite{Lieber:1992}.

\textit{\uline{Patterning muscle-tendon units onto backbone.}}
We pattern the sixteen muscle-tendon units onto the elastic backbone. Units are arranged in four layers. Each layer consists of four units, organized in two agonist-antagonist pairs oriented orthogonal to each other. Neighboring pairs overlap lengthwise by 50\% while being rotated 45\textdegree\ relative to each other to avoid intersecting. Muscle-tendon units are arranged on the elastic backbone's surface such that the muscle-tendon unit's center-line lies one (backbone) radius away from the backbone's center-line. 
% This causes the muscle-tendon unit to partially intersect the backbone, which is accounted for by reducing the density of the backbone proportional to the volume of the intersecting muscle-tendon unit, resulting in the backbone being modeled as if sections were carved out to fit the inner half of each unit. 
The arm in its rest configuration presents no muscle activation or residual stresses.

To join the muscle-tendon units to the backbone, we define zero-displacement boundary conditions at the ends of the muscle-tendon unit relative to point on the outer surface of the backbone to model to first order the enthesis point where tendon and bone join together. The location of these connections a radial distance from the backbone's center-line then generates a bending torque on the backbone then the muscle-tendon unit linear shortens when activated. 
Beyond these primary connections, we also implement distributed displacement-force boundary conditions to lightly adhere the muscle-tendon unit to the backbone during bending motions \cite{Tekinalp:2024, Zhang:2019}, analogous to the effect of connective fascia tissue that helps maintain muscle organization. 
% We define the connection between the outer surface of the elastic backbone and the center-line of the muscle-tendon unit. 

\subsubsection*{3D tracking environment} 
The muscular arm defined above is challenged to track a target moving in 3D space along to a smoothly varying, random trajectory. The target trajectory $\mathbf{\bar{x}}_t (t) = [\bar{x}_1, \bar{x}_2, \bar{x}_3]$ is defined at the beginning of each episode with each spatial component of the trajectory generated according to 
\begin{equation}
    \label{eq:trajectory}
    \bar{x}_j(t) = b_j \sin(f_{j,1} 2 \pi  t) \sin(f_{j,2} 2 \pi  t) \sin(f_{j,3} 2 \pi  t) \quad \forall j \in [1..3]
\end{equation} 
where $f_{j,i} \in \mathbb{R}$ is a randomly selected constant from the interval $[0.5, 1.0]$ drawn at the beginning of each episode and $b_j$ is a velocity scaling constant, whose sign is also randomly selected each episode. This structure of Eq. \ref{eq:trajectory} results in a smoothly varying trajectory of the target while the random selection of $f_{j,i}$ each episode renders it not possible for the controller to simply memorize a target trajectory. 
The state and muscle activations of the arm are updated at a 4 Hz control frequency ($\Delta t$ = 250 ms), leading to a mapping between the physical time $t$ of the simulation environment and the control time step number $n$ of $n = [4 t]$.
At each timestep, muscle activations are directly mapped from the action space $\textbf{a}^n = [a_i] \in \mathbb{R}^{16}$, $a_i \in [-\infty, \infty]$ to an individual muscle activation level $\alpha_i(t) = (\tanh(a_i(t))+1)/2 \in [0, 1]$ which is applied for 0.25 seconds (until the next control time step). The arm's state consists of the current 3D target location $\mathbf{\bar{x}}_t (t)$ and the current averaged normalized curvature of the arm ${\bm{K}}_i(t)$ over $i=4$ equidistant intervals in the rod's local normal and binormal directions 
\begin{equation}
    % {\bm{K}}_i(t) = \frac{L}{2 \pi}\frac{\int_{d_i}^{d_{i+1}} \bm{\kappa}(s,t) ds}{\int_{d_i}^{d_{i+1}} ds} \quad \forall i \in [0..3]
    {\bm{K}}_i(t) = \frac{L}{2 \pi}\frac{\int_{d_i}^{d_{i+1}} \bm{\kappa}(s,t) ds}{d_{i+1} - d_i} \quad \forall i \in [0..3]
\end{equation}
where $\bm{\kappa}(s,t) \in \mathbb{R}^2$ is the local normal and binormal curvature of the arm, $d = \{k L/4 | k \in [0..4]\}$, and $L$ is the length of the arm. This results in a state $\textbf{s}^n = [\mathbf{\bar{x}}_t (t), {\bm{K}}(t)] \in \mathbb{R}^{3+8}$. The reward is the square of the Euclidean distance between the tip of the arm $\bar{\mathbf{x}}(L, t)$ and the 3D moving target $\mathbf{\bar{x}}_t (t)$ integrated over the control time step
\begin{equation}
    r_n = \int_{t}^{t+\Delta t} ||\bar{\mathbf{x}}(L, t') - \mathbf{\bar{x}}_t (t') ||^2 dt' 
\end{equation}
The integration of the reward over the time step ensures the tip of the arm continuously follows the target trajectory over the entire time step. Each episode is terminated after 100 seconds.

\subsubsection*{Unstructured nest environment} 
In Fig. \ref{fig:4}, the arm is challenged to reach through an unstructured nest of rigid cylindrical obstacles towards a stationary target. Here, the target and nest are stationary and fixed in the same location for all episodes. Between the target and the arm's initial configuration, an unstructured nest of cylindrical obstacles is placed, which the arm must navigate through to reach the target on the other side. The location of the obstacles is the same for all episodes and each episode is terminated after 5 seconds. External contact forces between the arm and the nest are implemented (described in SI) as well as contact between the arm and the floor, allowing the arm to interact with its external environment. The state, action, and reward of this environment are the same as with the 3D tracking environment, but here successful control of the neuromuscular arm requires addressing a more complex physical environment by learning to interact with and account for the presence of solid obstacles that restrict reaching.  

\subsection*{Neural reservoir formulation} 
We consider two formulations of a neural reservoir: artificial neuron reservoirs and spiking neuron reservoirs. Artificial neuron reservoirs consist of a discrete-time, recurrent network of artificial neurons typically used in machine learning applications while spiking neuron reservoirs utilize a continuous-time, spike-based approach.

\subsubsection*{Neural reservoir with artificial neurons}
Neural reservoirs using artificial neurons are implemented using an Echo State Network (ESN) approach \cite{Jaeger:2001}. The reservoir consists of a large pool of $n$ sparsely and recurrently connected artificial neurons. Recurrent connections within the reservoir are represented using the reservoir connection matrix $W_{r}\in \mathbb{R}^{n\times n}$ while connections into the reservoir from the input state (of size $s$) are represented using the input connection matrix $W_{i}\in \mathbb{R}^{n\times s}$. 
The state of the reservoir $\mathbf{u}_{r}^n$ at time step $n$ is then 
\begin{equation}
\label{eq:ESN}
    \mathbf{u}_{r}^n = f(W_{i} \mathbf{s}^n + W_{r}\mathbf{u}_{r}^{n-1})
\end{equation}
where $f(\cdot)$ is a non-linear activation function. Here, a hyperbolic tangent function (TanH) is used. Notably, the connection weight matrices are fixed at initialization. No adjustment to their weights is made during the learning process. Thus, their proper initialization is critical to the information processing capability of the reservoir. Following the ESN approach, both the reservoir connection matrix $W_r$ and the input connection matrix $W_i$ are randomly initialized from a normal distribution with zero mean, a standard deviation of 0.50, and a sparse connection density $\rho$ of 0.10. The reservoir connection matrix is then normalized based on its spectral radius $\lambda$ (the magnitude of its largest eigenvector) to ensure the echo state property, which is that information from initial conditions will asymptotically disappear over time. To achieve the echo state property it is generally sufficient (though not necessary) to enforce $\lambda < 1$ \cite{Jaeger:2001}.

The mapping from the reservoir state $ \mathbf{u}_{r}^n$ to the output (in this case action, with size $j$) is determined according to the linear output mapping $W_o \in \mathbb{R}^{j\times n}$ according to
\begin{equation}
    a^n = W_o \mathbf{u}_r^n
\end{equation} 
The linear map $W_o$ can be learned using a variety of different algorithms.
For the control policy, the unsupervised reinforcement learning Proximal Policy Optimization (PPO) algorithm was used while for the parallel output maps of Fig. \ref{fig:3}, supervised learning algorithms such as ridge regression were used.

\subsubsection*{Neural reservoir with spiking neurons} 
To implement a neural reservoir using spiking neurons, we use leaky-integrate and fire (LIF) neurons \cite{Abbott:1999} whose dynamics are governed by the differential equation
\begin{align}
    \tau_m \frac{dv}{dt} = (v_{r} -v(t)) + RI(t)
\end{align}
where $\tau_m$ is the membrane constant of the RC circuit simulating voltage leakage, $R$ is the resistance, $I(t)$ is the input current, $v(t)$ is the output voltage, and $v_{r}$ is the reset membrane potential. When $v(t)$ exceeds some threshold voltage level ($v(t) > v_{t}$), a spike in voltage occurs at time $t$ followed by the neuron returning to $v_{r}$ which it is held at for some refractory time $\tau_r$. 

We begin by converting the discreet time, continuous state input of the soft muscular arm to a continuous time, spike-based input compatible with a spiking neural reservoir. Each state variable is presented as current into two LIF input neurons for a time of $\Delta$ (which we decouple from the control time $\Delta t$). 
These two neurons have opposite tuning curves such that one will positively rate encode increasing values while the other will rate encode decreasing values. The resulting spike trains are multiplied by input weights $W_i$ and fed into the spiking reservoir. The input weights $W_i$ are randomly drawn from a uniform distribution between -1 and 1. The spiking neural reservoir consists of a set of $N$ LIF neurons recurrently connected by a sparse set of weights $W_r$. The recurrent connection weights have a density of 0.1 and are drawn from the normal distribution $\mathcal{N}(0.0,0.5^2)$. LIF neuron properties such as refractory period and membrane voltage decay as well as reservoir connection properties were optimized using a CMA evolutionary search algorithm \cite{Hansen:2003} as detailed in the SI.
For each control time step, the number of spikes over the time interval $\Delta$ is counted for each reservoir neuron. This spike count is then treated as the output of the reservoir for that control time step. This approach then enables the same RL algorithms as used to train the artificial neural reservoir to be employed to learn the output weights $W_o$ to control the arm.  

This neural reservoir setup was deployed on Intel's Kapoho Bay device \cite{Davies:2018}, which contains 2 Loihi chips. The reservoir was defined using the Nengo Python package \cite{Bekolay:2014} and its NengoLoihi extension. The input spike conversion layer and output spike counting layer are implemented off-chip, with the communication between the CPU host and the Loihi chip occurring only via spike communication.

\subsubsection*{Energy efficiency}
The energy usage of neural reservoirs consisting of either artificial neurons or spiking neurons was analyzed by measuring the average energy consumption for each inference, or control time step. For a neural reservoir with artificial neurons, this refers to a single evaluation of the system defined in Eq. \ref{eq:ESN} while for spiking neurons it refers to the presentation of the input state for a time $\Delta$. The energy consumption of the artificial neural reservoir was measured using the pyRAPL Python package on an Intel Xeon W-2265 CPU to identify the running power with idle power subtracted. Energy usage on Loihi was measured using utilities provided by Intel's NxSDK API for Loihi to similarly measure running power.

\subsection*{Reinforcement learning integration}
To learn the mapping from the neuromuscular arm's state to the action required to control it, reinforcement learning (RL) is used. We employ the state-of-the art, model-free, on-policy Proximal Policy Optimization (PPO) algorithm to optimize $W_o$  \cite{Schulman:2017} based on the open-source Stable Baselines 3 implementation \cite{stable-baselines3}. When used with feed-forward and LSTM network architectures, the environment state $\mathbf{s}^n$ was used as the input state into the artificial neural network in the usual manner. When integrated with the neural reservoir, the current reservoir state $\mathbf{u}_r^n$ is used as the input state to the PPO algorithm, with the PPO algorithm then tasked with learning the single layer, linear mapping $W_{o}$ from the reservoir state to the optimal muscle action. When a neural reservoir consisting of artificial neurons is used, the reservoir state is directly used as the input into the PPO algorithm while for the spiking neural reservoir, output spikes are accumulated and counted over each control time step to determine an analogous reservoir state that can be then directly used by the PPO algorithm.  

\subsubsection*{Training and evaluation}
All network architectures were trained using the same approach. For the 3D tracking case (Figs. \ref{fig:1}, \ref{fig:2}, \ref{fig:4}b,c), the policy was trained using 120 parallel environments. The policy was updated every 4800 time steps (1200 seconds at 4 Hz), resulting in each environment collecting ten seconds of data between updates. Each environment was reset every 100 seconds with a new trajectory generated and the arm initiated in a straight configuration. Policies were trained for either two or four million time steps and five separate policies with random initial seeds were trained for each case. Trained policies were evaluated on new, unseen trajectories for up to 100 seconds (400 time steps). Hyperparameter tuning for all algorithms was performed as detailed in the SI. All training and hyperparameter tuning was performed on the Pittsburgh Supercomputing Center's Bridges 2 cluster.

For the spiking neural reservoir, only one Kapaho Bay device was available, so NengoLoihi's emulation environment, which emulates Loihi hardware on CPUs, was used to train using the same 120 parallel environment approach for the 3D tracking case (Fig. \ref{fig:4}b,c). 
For the unstructured obstacle case (Fig. \ref{fig:4}e,f), 120 parallel environments were also used (using the NengoLoihi emulator), resulting in each environment having a five-second episode, however, environments were then reset after each episode. All training using the Loihi emulation environment was performed on the Pittsburgh Supercomputing Center's Bridges 2 cluster.
Trained policies were then evaluated on Loihi hardware to determine performance and energy usage. All evaluation and visualization results reported in Fig. \ref{fig:4} and in the SI videos were run on Loihi hardware.

\subsection*{Kinetic and bending energy measurements}
To report the kinetic and bending energy measurements in Figure \ref{fig:2}b,c, the average kinetic and bending energy of the arm's backbone over an episode of length $T$ was measured as
\begin{equation}
    \bar{e}_i = \frac{\int_0^T e_i(t) \ dt}{\int_0^T dt}
\end{equation}
where $e_i \in \{e_k, e_b\}$. Here $e_k$ is the kinetic energy of the backbone and $e_b$ is the elastic bending energy of the rod. 
The kinetic energy of the rod at each time is 
\begin{equation}
    {e}_k(t) = \frac{1}{2}\int_0^L \rho A (\bm{v} \cdot \bm{v})  \ ds
\end{equation}
where $\rho$ is the local density, $A$ is the local cross-sectional area, and  $\bm{v}$ is the local velocity.
The bending energy of the rod at each time is 
\begin{equation}
    {e}_b(t) = \frac{1}{2}\int_0^L \bm{\kappa} \cdot \bm{\tau}\ ds
\end{equation}
where $\bm{\kappa}$ is the local curvature of the rod, $\bm{\tau} = \mathbf{B} \bm{\kappa} $ is the local internal elastic torque vector, and $\mathbf{B}$ is the rod's bend/twist stiffness matrix.
Arm dynamics from trained policies were recorded for 100 unique episodes (100 seconds each) and the average energy of each episode was calculated to determine the distribution of values reported in Fig. \ref{fig:2}b.

\subsection*{Training of parallel maps} 
Figure \ref{fig:3} shows the use of parallel maps to increase the information extracted from the neural reservoir. 
To generate these parallel maps, an RL policy trained to track a moving target was recorded for 100 seconds over 400 different target trajectories.
The state, action, reservoir state, and environment were recorded. These data were then used as the training data (using an 80/20 training/testing data split) for parallel maps using a supervised learning method via ridge regression. For target correction (Fig. \ref{fig:3}d), the parallel map was evaluated every time step to predict the 3D position the target would be one time step in the future and used in place of the target's true position in the state vector as required.

\end{document}